\documentclass[journal]{IEEEtran}
\usepackage{cite}
\usepackage{amsmath}
\usepackage{newtxmath}
\usepackage{graphicx}
\usepackage{algorithm,algorithmic}
\usepackage{bm}
\usepackage{booktabs}
\usepackage{siunitx}
\usepackage{hyperref}
\usepackage{cancel}
\usepackage{textcomp}
\def\BibTeX{{\rm B\kern-.05em{\sc i\kern-.025em b}\kern-.08em
    T\kern-.1667em\lower.7ex\hbox{E}\kern-.125emX}}

\newcommand{\x}{\bm{x}}
\newcommand{\y}{\bm{y}}

\renewcommand{\vv}{\bm{v}}
\newcommand{\A}{\mathcal{A}}
\newcommand{\Hm}{\bm{H}}
\newcommand{\Ne}{N_{\mathrm{elem}}}
\newcommand{\Nb}{N_{\mathrm{beams}}}

\begin{document}
\title{Deep Bayesian REFoCUS}

\author{Simon W. Penninga
and Ruud J. G. van Sloun
\thanks{This work was supported by the European Research Council (ERC) under the ERC Starting Grant US-ACT
(grant no.~101077368). \textit{(Correspondence: Simon W. Penninga.)}}
\thanks{S. W. Penninga and R. J. G. van Sloun are with the Department of Electrical Engineering, Eindhoven University of Technology,
Eindhoven, The Netherlands (e-mail: s.w.penninga@tue.nl;
r.j.g.v.sloun@tue.nl).}}

\maketitle

\begin{abstract}
In this work we formulate ultrasound multistatic recovery from arbitrary transmit sequences as a Bayesian inference problem. To that end, we train a deep generative prior on multistatic data sets to tackle the rank-deficient regime in which classical linear REFoCUS decoders fail. This appproach, which we term Deep Bayesian REFoCUS, outperforms the linear baselines for all regimes of rank-deficiency and noise levels, and regresses to linear decoding when inversion is exact. The model also expresses uncertainty in the null space of the acquisitions, whereas the linear REFoCUS decoders only provide point estimates. Finally, we analyze the impact of distribution shift between simulation and in-vivo acquisitions, showing remarkable generalization ability without any fine-tuning or adaptation.
\end{abstract}

\begin{IEEEkeywords}
ultrasound imaging, multistatic recovery, full matrix capture,
retrospective transmit beamforming, diffusion models,
inverse problems, posterior sampling.
\end{IEEEkeywords}

\section{Introduction}
\label{sec:introduction}

\IEEEPARstart{T}{he} diagnostic quality of a B-mode ultrasound image is strongly influenced by its transmit sequence. Classical line-by-line imaging focuses a series of transmits at spatial focal points, sweeping across the Field Of View (FOV). This achieves relatively high contrast and good resolution at the cost of frame rate due to the large number of transmits needed to cover the view. Ultrafast sequences such as plane-wave and diverging-wave transmission raise the frame rate by orders of magnitude but suffer from reduced contrast, worse lateral resolution, and a lower signal-to-noise ratio (SNR)~\cite{montaldo2009coherent,
tanter2014ultrafast}. Coherent compounding of multiple steered transmits can recover part of the loss~\cite{montaldo2009coherent,papadacci2014high}, but the underlying trade-off between frame rate, SNR, and focusing quality remains. Moreover, coherent compounding of unfocused transmits that cover a large FOV is impacted by tissue motion across the transmit sequence, and generating sufficient tissue harmonics for harmonic imaging is more difficult.

In the linear regime, the multistatic dataset (recordings at every receive element for every transmit element firing individually) enables retrospective transmit and receive focusing with the full array entirely computationally, via linear combinations of timed single-element firings. However, acquiring a high-quality multistatic data set is challenging in practice, since sequential single-element firing severely limits SNR and frame rate~\cite{jensen2006synthetic}.

With Retrospective Encoding For Conventional Ultrasound Sequences (REFoCUS)~\cite{bottenus2018refocus}, the multistatic data set can instead be recovered from arbitrary transmit sequences. This is done by inverting the linear encoding of single-element responses. Recording the channel data of a conventional full-aperture sequence and applying the
adjoint of the encoding (a re-delay-and-sum across transmit events) recovers an estimate of the complete data set but with full-aperture transmit SNR. This approach was later reformulated in the frequency domain and stabilized with Tikhonov regularization, extending recovery to steered and walking-aperture sequences~\cite{ali2020extending}. However, both decoding methods yield estimates that are ultimately linear in the measurements, and exact inversion only holds for a set of noiseless transmits that is full rank. 

In this paper we introduce Deep Bayesian REFoCUS (DB-REFoCUS) to tackle the rank deficiency constraint. In the rank-deficient regime, a learned prior can exploit statistical dependencies between observed and unobserved components to infer a distribution over plausible multistatic sets. Real multistatic data sets are highly structured and spatially coherent across the elements, and we train a deep generative model to capture this structure and to recover the multistatic data set through posterior sampling.

\begin{figure*}[th]
    \centering
    \includegraphics[width=\linewidth]{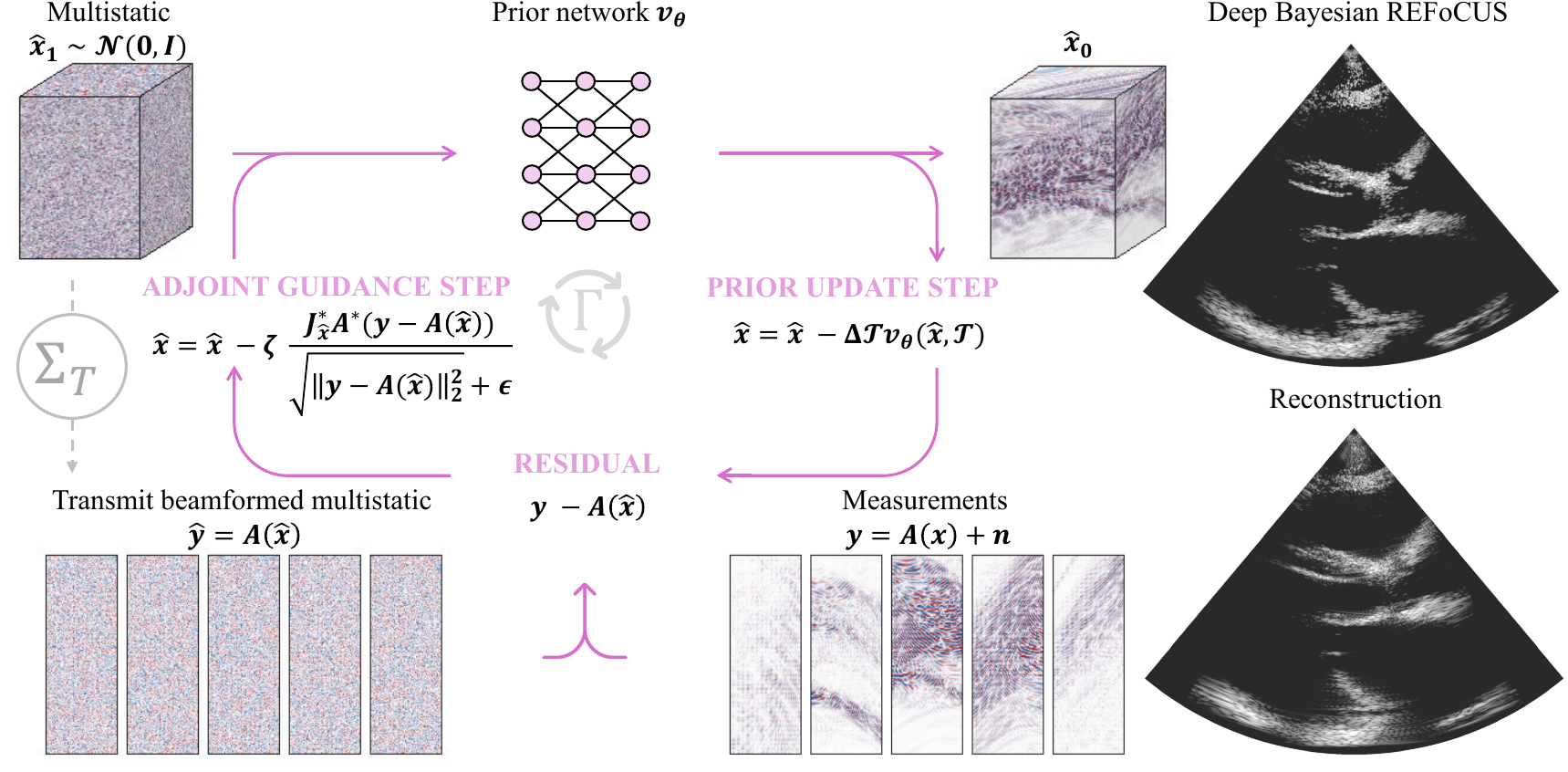}
    \caption{Schematic of the Deep Bayesian REFoCUS sampling process. A multistatic data set is initialized with random Gaussian noise. The diffusion process is guided by the residual of the measurements, which we obtain through transmit beamforming. The neural network steps introduce the multistatic prior statistics, and the adjoint guidance steps transport the generation process towards the measurements.}
    \label{fig:overview}
\end{figure*}

The recent introduction of diffusion and flow-matching models~\cite{ho2020denoising,
song2021scorebased,lipman2023flow,liu2023flow} has made learning large generative priors (e.g. images and videos) feasible, and posterior-sampling schemes~\cite{song2022solving,kawar2022ddrm,
song2023pseudoinverse}
such as Diffusion Posterior Sampling (DPS)~\cite{chung2023diffusion} have demonstrated the ability to solve ill-posed inverse problems. Here we show that these techniques can also be used to learn priors over raw ultrasound channel data. We link DPS guidance with the classical adjoint-based REFoCUS, and show that it can be directly used with any ultrasound sequence (without training for specific sequences), only requiring the specific acquisition operator at inference. 
Repeated sampling from the posterior gives us an ensemble whose spread represents epistemic uncertainty, which correlates strongly with model error.
For an overview of DB-REFoCUS see Fig.~\ref{fig:overview}. 

In the remainder of this paper we review related work and derive a Bayesian representation of REFoCUS. We train a generative prior on idealized simulated multistatic data sets and evaluate on in vivo cardiac recordings of the Parasternal Long Axis (PLAX) view for various transmit counts.

\section{Related Work}
\label{sec:related}
Classical Synthetic Transmit Aperture (STA) imaging fires one element at a
time and recovers the multistatic data set with single-element transmit SNR~\cite{karaman1995synthetic,jensen2006synthetic}. Much of the subsequent work in multistatic imaging is aimed at improving on this single-element SNR. Temporal coding methods replace the short pulse with a modulated waveform that is compressed on
receive~\cite{odonnell1992coded,misaridis2005modulated}, multi-element transmits emulate a spherical wave from a defocused subaperture or from a virtual source behind the array~\cite{gammelmark2003multielement,
bae2000study,nikolov2002virtual}, and spatial encodings fire the full
aperture with orthogonal (Hadamard) sign patterns whose decoding returns the per-element responses~\cite{chiao1997sparse,tiran2015multiplane}. The codes are chosen specifically such that decoding is orthogonal and trivial, which creates a limitation for transmit design.

Retrospective transmit focusing was initially done by deconvolving the fixed transmit ``lens'' to synthesize an in-focus image outside of the focal zone, using least squares optimization~\cite{freeman1995retrospective}.
REFoCUS generalized the retrospective focusing to the complete data set and removed the restriction on transmit design, making any
transmit sequence decodable with the adjoint~\cite{bottenus2018refocus}, ramp-filtered adjoint, and Tikhonov-regularized least
squares~\cite{ali2020extending}.
However, when the number of transmit events is smaller than the number of transducer elements, the per-frequency encoding matrix necessarily has a nontrivial null space. Consequently, the multistatic responses cannot be uniquely determined from the measurements alone without additional assumptions/priors.

Existing approaches address this nonuniqueness through different assumptions: compressed-sensing methods exploit sparsity~\cite{liu2017cssta}, whereas minimum-norm least-squares methods select the solution with zero null-space component~\cite{zhang2022partialhadamard}. With the advent of deep learning, directly learning non-linear decoders became opportune. In \cite{zhang2023deepnullspace}, a convolutional network trained in the null space of the Hadamard encoding regresses the residual, and in \cite{chen2021apodnet} transmit apodizations are learned jointly with a decoding network.

These approaches are discriminative in nature: a network is trained to regress a target image, apodization, or channel estimate using some cost function, for a fixed acquisition scheme, and any new transmit setting requires retraining.
Moreover, a non-linear point regressor does not express the spread of the null space and diversity of plausible solutions.

Instead, we learn a generative prior over complete data sets and supply the encoding as a differentiable operator only at inference, retaining universality for a particular anatomical domain and bandwidth. 
Recovery is done through posterior \emph{sampling}: the null space is populated by draws from the conditional distribution of the unobserved components given the measurements, and repeated
sampling shows which part of the transmit acquisition is under-determined. 

\section{Methods}
\label{sec:methods}

\subsection{The multistatic data set and encodings}
\label{sec:problem}
We adhere to the original notation of the REFoCUS model~\cite{ali2020extending}.
Consider an array of $\Ne$ elements and let $u_{TR}(t)$ denote the echo signal
received at time $t$ at element $R \in \{1, \dots, \Ne\}$ after firing transmit element $T \in \{1, \dots, \Ne\}$. The collection
of these responses over all element pairs $(T, R)$ is the multistatic data
set. Because acoustic propagation is linear, the channel data produced by any transmit event can be seen as a superposition of the single-element responses. A transmit event $n$ that
fires $\Ne$ elements with apodization weights $w_{nT}$ and delays $\tau_{nT}$
records
\begin{equation}
  s_{nR}(t)
  = \sum_{T=1}^{\Ne} w_{nT}\, u_{TR}(t - \tau_{nT}) + n_{nR}(t),
  \label{eq:encoding}
\end{equation}
where $n_{nR}(t)$ is measurement noise. Taking the Fourier transform along the
fast-time axis turns the delays into phase ramps and decouples the problem per
temporal frequency $f$ and receive element $R$:
\begin{equation}
  S_{nR}(f) = \sum_{T=1}^{\Ne} H_{nT}(f)\, U_{TR}(f),
  \label{eq:encodingfreq}
\end{equation}
with the encoding-matrix entries
\begin{equation}
  H_{nT}(f) = w_{nT}\, e^{-j 2\pi f \tau_{nT}}.
  \label{eq:encodingmatrix}
\end{equation}
Stacking the $\Nb$ recorded events per frequency bin and receive
channel gives $\bm{S}_{R}(f) = \Hm(f)\, \bm{U}_{R}(f)$ with
$\Hm(f) \in \mathbb{C}^{\Nb \times \Ne}$ defined as

\begin{equation}
  \Hm(f) =
  \begin{bmatrix}
    H_{11}(f) & \cdots & H_{1\Ne}(f) \\
    \vdots & \ddots & \vdots \\
    H_{\Nb 1}(f) & \cdots & H_{\Nb\Ne}(f)
  \end{bmatrix}.
  \label{eq:encodingmatrixfull}
\end{equation}

Accurate recovery of $\bm{U}_{R}(f)$ is then dependent on the pseudo-inverse $\Hm^\dagger$.

Initially, the underlying multistatic data set was estimated using the adjoint of the encoding
\begin{equation}
  \widehat{\bm{U}}_{R}(f) = \Hm^{*}(f)\, \bm{S}_{R}(f),
  \label{eq:adjoint}
\end{equation}
where $(\cdot)^{*}$ denotes the conjugate transpose. In the time domain this
equals a re-delay-and-sum across transmit events.
The adjoint's nonuniform frequency weighting was later corrected with a ramp filter and an additional recovery method was proposed with the Tikhonov-regularized inverse
\begin{equation}
  \widehat{\bm{U}}_{R}(f)
  = \left[\Hm^{*}(f)\, \Hm(f)
    + \left(\gamma\, \sigma_{\max}(f)\right)^{2} \bm{I}\right]^{-1}
    \Hm^{*}(f)\, \bm{S}_{R}(f),
  \label{eq:tikhonov}
\end{equation}
where $\sigma_{\max}(f)$ is the largest singular value of $\Hm(f)$ and
$\gamma$ a regularization parameter~\cite{ali2020extending}. The regularized inverse
equalizes the encoding where it is invertible and suppresses noise
amplification where it is not. Both estimators are linear, and their accuracy is determined by the rank of the $\Hm(f)$-matrix. For $\Nb < \Ne$ every bin has an at least $(\Ne - \Nb)$-dimensional null
space.

\subsection{Bayesian REFoCUS}
\label{sec:bayesian}
To impose a statistical distribution we rewrite the multistatic data set as a sampleable random variable $\x$ in the time domain, with the set of measurements $\{s_{nR}(t)\}$ as $\y$.
The acquisition is formulated as 
\begin{equation}
  \y = \A(\x) + \bm{n},
  \label{eq:forward}
\end{equation}
with $\bm{n}$ complex Gaussian noise with diagonal covariance.
The transmit beamforming operator $\A$ follows \eqref{eq:encodingfreq} through
\begin{equation}
    \A(\x) = \mathcal{F}^{-1}(\Hm\mathcal{F}(\x))
\end{equation}
and is still defined by $w_{nT}$ and $\tau_{nT}$,
with $\mathcal{F}$ the fast Fourier transform across the fast-time axis. The posterior multistatic data set $\x$, conditioned on measurements $\y$, is expressed through Bayes' rule
\begin{equation}
p(\x \mid \y) = \frac{p(\y \mid \x)p(\x)}{p(\y)}.
    \label{eq:bayes}
\end{equation}

We model the prior $p(\x)$ with a continuous-time flow-matching generative model~\cite{lipman2023flow,liu2023flow,albergo2023stochastic}. This type of score model regresses the velocity of a linear interpolation between data samples $\x_0 \sim p(\x)$ and Gaussian noise $\bm{\epsilon} \sim \mathcal{N}(\bm{0}, \bm{I})$.
The relation is defined as 
\begin{equation}
  \x_{\mathcal{T}} = (1 - \mathcal{T})\,\x_{0} + \mathcal{T}\,\bm{\epsilon}, \qquad \mathcal{T} \in [0, 1],
  \label{eq:interp}
\end{equation}
with velocity
$\partial\x_\mathcal{T}/\partial\mathcal{T}=\vv = \bm{\epsilon} - \x_{0}$. A network
$\vv_{\theta}(\x_{\mathcal{T}}, \mathcal{T})$, parameterized by weights $\theta$, is trained to minimize the flow-matching objective
\begin{equation}
  \mathcal{L}(\theta)
  = \mathbb{E}_{\x_{0}, \bm{\epsilon}, \mathcal{T}}
    \left\| \vv_{\theta}(\x_{\mathcal{T}}, \mathcal{T}) - (\bm{\epsilon} - \x_{0}) \right\|_{2}^{2}.
  \label{eq:fmloss}
\end{equation}
With a model for the velocity field, the noising process can be inverted to obtain an unconditional multistatic data set estimate
\begin{equation}
  \hat{\x}_{0}(\x_{\mathcal{T}}, \mathcal{T}) = \x_{\mathcal{T}} - \mathcal{T}\, \vv_{\theta}(\x_{\mathcal{T}}, \mathcal{T}).
  \label{eq:denoiser}
\end{equation}
The posterior of a score-matching model is obtained through
\begin{equation}
    \log p(\x_\mathcal{T} \mid \y) = \log p(\y \mid \x_\mathcal{T}) + \log p(\x_\mathcal{T}) - \log p(\y)
\end{equation}
such that
\begin{equation}
\begin{aligned}
\nabla&_{\x_\mathcal{T}} \log p(\x_\mathcal{T} \mid \y)
\\ &= \nabla_{\x_\mathcal{T}} \log p(\y \mid \x_\mathcal{T})
 + \nabla_{\x_\mathcal{T}} \log p(\x_\mathcal{T}) - \nabla_{\x_\mathcal{T}} \log p(\y)\\
&\approx \nabla_{\x_\mathcal{T}}
\log p\!\left(\y \mid \hat{\x}_0\right)
 + \nabla_{\x_\mathcal{T}} \log p(\x_\mathcal{T}).
\end{aligned}
\label{eq:score}
\end{equation}
The final expression uses $\nabla_{\x_\mathcal{T}} \log p(\y)=0$ and the assumption $p(\y \mid \x_{\mathcal{T}}) \approx p(\y \mid \hat{\x}_{0})$, following DPS~\cite{chung2023diffusion}.
With the Gaussian noise model in \eqref{eq:forward} the likelihood becomes
\begin{equation}
\begin{aligned}
\nabla_{\x_\mathcal{T}} \log p(\y \mid \hat{\x}_0)
&=
-\frac{1}{2\sigma_{\y}^2}
\nabla_{\x_\mathcal{T}}
\left\|
\y-\A\!\left(\hat{\x}_0(\x_\mathcal{T}, \mathcal{T})\right)
\right\|_2^2
\\
&=
\frac{1}{\sigma_{\y}^2}
J_{\hat{\x}_0}^*
\A^*
\left(\y-
\A\!\left(\hat{\x}_0(\x_\mathcal{T}, \mathcal{T})\right)
\right),
\end{aligned}
\end{equation}

with $J_{\hat{\x}_0}^*$ the adjoint Jacobian of the clean estimate. Note that this likelihood term contains the adjoint of the transmit encoding $\A^* = \mathcal{F}^{-1}\Hm^*\mathcal{F}(\cdot)$, operating on the residual of the measurements. This likelihood term contains the adjoint of the transmit encoding, which backprojects the measurement residual. It therefore uses the same adjoint operator as conventional REFoCUS for measurement consistency.

The posterior samples can be separated into two terms per frequency bin
\begin{equation}
  \widehat{\bm{U}}_{R}(f)
  \approx \underbrace{\Hm^\dagger(f)\, \bm{S}_{R}(f)}_{\text{measurement}}
  + \underbrace{\left(\bm{I} - \Hm^\dagger(f)\, \Hm(f)\right) \widehat{\bm{U}}_{R}(f)}_{\text{null space distributed by prior}}.
  \label{eq:decomp}
\end{equation}
The decomposition is not exact, as the reconstruction is not exactly measurement-consistent (the prior is not prevented from contributing to the part of the multistatic data set that is covered by measurements). This design choice is deliberate, since 
the measurements are noisy and should not be reproduced exactly. 
Instead, the guidance term is normalized and weighted by a parameter $\zeta$ in discrete diffusion time, resulting in the update step
\begin{equation}
  \x_{\mathcal{T}+1} = \x_\mathcal{T} - \zeta\frac{\nabla_{\x_\mathcal{T}}\log p(\y\mid\hat{\x}_0)}{\sqrt{\log p(\y\mid\hat{\x}_0)} + \varepsilon}\,
  \label{eq:dpsstep}
\end{equation}
with $\varepsilon$ some small stability constant. 
The complete multistatic recovery is shown in Algorithm~\ref{alg:dps}.
\begin{algorithm}[ht]
\caption{Deep Bayesian REFoCUS}
\label{alg:dps}
\begin{algorithmic}[1]
\STATE \textbf{Input:} transmit beamforming operator $\A$, measurement $\y$, velocity
network $\vv_{\theta}$, step count $\Gamma$, number of posterior samples $K$, guidance strength scalar $\zeta$.
\STATE schedule $1 = \mathcal{T}_{0} > \mathcal{T}_{1} > \cdots > \mathcal{T}_{\Gamma} = 0$
\FOR{$k = 0$ \TO $K-1$} 
    \STATE $\x_{\mathcal{T}_0} \sim \mathcal{N}(\bm{0}, \bm{I})$
    \FOR{$\rho = 0$ \TO $\Gamma - 1$}
      \STATE $\hat{\x}_{0} \leftarrow \x_{\mathcal{T}_\rho} - \mathcal{T}_{\rho}\, \vv_{\theta}(\x_{\mathcal{T}_{\rho}}, \mathcal{T}_{\rho})$
      \STATE $r \leftarrow \y - \A(\hat{\x}_{0});\quad L \leftarrow ||r||^2_2$
      \STATE $\nabla_{\x_{\mathcal{T}_\rho}}L = J^*_{\hat{\x}_{0}}\A^*r$
      \STATE $\x_{\mathcal{T}_{\rho+1}} \leftarrow \x_{\mathcal{T}_{\rho}} + \dfrac{\zeta}{\sqrt{L} + \varepsilon}\, \nabla_{\x_{\mathcal{T}_\rho}}L$
      \STATE $\x_{\mathcal{T}_{\rho+1}} \leftarrow \x_{\mathcal{T}_{\rho+1}} + \vv_\theta(\x_{\mathcal{T}_\rho},\mathcal{T}_\rho)(\mathcal{T}_{\rho+1} - \mathcal{T}_\rho)$
    \ENDFOR
    \STATE $\x^k \leftarrow \x_{\mathcal{T}_\Gamma}$
\ENDFOR
\RETURN $\{\x^0, \dots, \x^{K-1}\}$ \hfill $\triangleright$ Posterior samples $\sim p(\x\mid\y)$
\end{algorithmic}
\end{algorithm}

\section{Experimental Setup}
\label{sec:experiments}
\subsection{Model prior}
\label{sec:expmodelprior}
\begin{figure}[th]
    \centering
    \includegraphics[width=\linewidth]{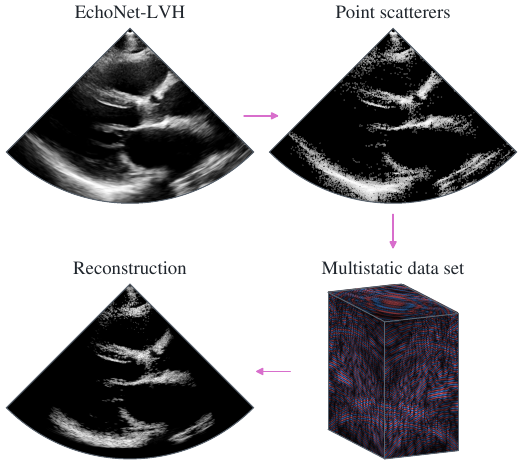}
    \caption{Multistatic data generation example for an EchoNet-LVH frame. Point scatterers are sampled based on pixel intensities, full-rank STA firings are simulated with zea. For reference, the multistatic data set is beamformed.}
    \label{fig:simulation}
\end{figure}

To train a generative prior we simulate a collection of multistatic data sets derived from real cardiac anatomy. B-mode images of the EchoNet-LVH data set~\cite{duffy2022highthroughput} are used (cardiac PLAX view), and point scatterers are sampled from pixel intensities of every frame. The scatterer positions are jittered and mapped to a lateral extent of
$\pm\SI{70.7}{\milli\meter}$ and a depth of 0--\SI{100}{\milli\meter}, creating a sector of $\pm45^\circ$. With zea~\cite{zea}, STA acquisitions are simulated for a Philips S5-1 probe with $\Ne=80$, \SI{0.25}{\milli\meter} pitch,
$f_{c} = \SI{4}{\mega\hertz}$, $f_{s} = \SI{12.5}{\mega\hertz}$, $c=$\SI{1540}{\meter\per\second}, $N_{\mathrm{ax}}=1536$, $\Nb=80$. The simulator is linear and operates in the frequency domain. It does not include depth-wise attenuation, measurement noise, sound speed variation, or out-of-plane clutter. Moreover, using processed B-mode images that include noise, aberration effects, and clutter to create a scattering density map introduces a distribution bias. Our claim is therefore not that this is a realistic simulation of a true multistatic dataset, but that it is an idealized representation that is still sufficiently rich and expressive to construct a useful generative prior for multistatic acquisitions of impinging wavesfronts originating from some anatomical structure. 
A schematic showing the data set generation process is shown in Fig.~\ref{fig:simulation}.
\begin{table}[th]
\centering
\caption{U-Net and training configuration.}
\label{tab:hparams}
\begin{tabular}{@{}lcccc@{}}
\toprule
Stage & Output (tx $\times$ ax $\times$ rx) & Channels & Blocks & Attn. \\
\midrule
Stem & $80 \times 512 \times 20$ & $96$  & ---  & --- \\
Encoder stage 1              & $80 \times 512 \times 20$ & $96$  & $4$  & --- \\
Encoder stage 2              & $40 \times 256 \times 10$ & $192$ & $3$  & \checkmark \\
Encoder stage 3              & $20 \times 128 \times 5$  & $384$ & $3$  & \checkmark \\
Bottleneck             & $10 \times 64  \times 5$  & $384$ & $3$  & \checkmark \\
Decoder stage 3              & $20 \times 128 \times 5$  & $384$ & $3$  & \checkmark \\
Decoder stage 2              & $40 \times 256 \times 10$ & $192$ & $3$  & \checkmark \\
Decoder stage 1              & $80 \times 512 \times 20$ & $96$  & $4$  & --- \\
Un-stem & $80 \times 512 \times 80$ & $2$ & ---  & --- \\
\midrule
\multicolumn{5}{@{}l}{\emph{Training}} \\
\midrule
Time sampling   & \multicolumn{4}{l@{}}{\texttt{logit-normal ($\mu = 0$, $s = 1$)}} \\
Optimizer       & \multicolumn{4}{l@{}}{\texttt{AdamW, weight decay $10^{-2}$}} \\
LR schedule     & \multicolumn{4}{l@{}}{\texttt{$10^{-4}$, cosine $\to 10^{-6}$}} \\
EMA momentum    & \multicolumn{4}{l@{}}{\texttt{0.9999}} \\
Batch / train steps  & \multicolumn{4}{l@{}}{\texttt{32 / 400,000}} \\
Parameters      & \multicolumn{4}{l@{}}{\texttt{146.7M}} \\
Precision       & \multicolumn{4}{l@{}}{\texttt{mixed bfloat16}} \\
Cluster         & \multicolumn{4}{l@{}}{\texttt{8x NVIDIA B200}} \\
Training time   & \multicolumn{4}{l@{}}{\texttt{240h}} \\
\bottomrule
\end{tabular}
\par\smallskip
{\footnotesize The (un-)stem layer is a single strided (transpose-)convolution. Attention (Attn.) denotes axis-factorized
self-attention with $8$ QK-normalized heads~\cite{vaswani2017attention,henry2020qknorm,dehghani2023vit22b}.\par
}
\end{table}

To reduce the computational cost of the model, multistatic data sets are compressed to IQ with $f_{\mathrm{demod}} = \SI{4}{\mega\hertz}$ and $3\times$ axial decimation, reducing model input to $\x \in \mathbb{R}^{\,80 \times 512 \times 80 \times 2}$ (transmit, fast-time, receive and complex axis, respectively). 
The multistatic tensors are normalized to zero mean and unit variance and clipped to $\pm 7$ standard deviations. The complete simulated corpus contains \num{107880} training samples and \num{11205} validation samples from held-out patients.

The architecture chosen for the velocity field model $\vv_{\theta}$ is a 3D convolutional U-Net~\cite{ronneberger2015unet}, built with residual blocks~\cite{he2016deep}. 
The network is trained with the
flow-matching objective \eqref{eq:fmloss}, with an XLA-compiled TensorFlow~\cite{abadi2016tensorflow} forward pass. An Exponential Moving Average (EMA) of the weights is used for all sampling and evaluation. All architecture and training parameters can be found in Table~\ref{tab:hparams}.

\subsection{Evaluation}
We compare DB-REFoCUS with the adjoint REFoCUS~\cite{bottenus2018refocus}, the ramp-filtered adjoint REFoCUS, and the Tikhonov-regularized inverse~\cite{ali2020extending}. For the Tikhonov method we perform an oracle grid search for $\gamma \in [10^{-4}, 10^1]$ for every data point to achieve a performance upper bound, sweeping 26 log-spaced points and selecting the best-performing result. For DB-REFoCUS we use $\Gamma=1000$ steps and report performance of the posterior mean with $K=8$ samples.

Multistatic recovery performance is measured in the channel domain with gain-compensated complex correlation~\cite{ali2020extending} and Normalized Mean-Squared Error (NMSE). In the image domain, we compare recovery with Peak Signal-to-Noise Ratio (PSNR) and the Structural Similarity Index Measure (SSIM). 
The multistatic data sets are converted to images through delay-and-sum beamforming with pressure-field weighting, envelope detection, 99.99th-percentile normalization and log compression, and are displayed with a fixed \SI{50}{dB} dynamic range.

We evaluate on six transmit families that are synthesized from the multistatic ground truth for simulation data, and recorded with a physical probe for the in vivo experiments. The following families are selected to cover a broad range of conditioning for the inverse problem:
\begin{itemize}
\item \emph{Focused sector beams (FOC)}: full-aperture transmits focused at a radius of \SI{60}{\milli\meter} ($\approx$F/3), steered across the sector. For in vivo experiments both fundamental and harmonic imaging modes are used.
\item \emph{Wide beams (WD)}: similar to FOC, but a focal radius of \SI{150}{\milli\meter}. Both fundamental and harmonic imaging modes for in vivo experiments.
\item \emph{Plane waves (PW)}: full-aperture steered waves with linear transmit delays.
\item \emph{Diverging waves (DW)}: steered virtual point sources with a radius of \SI{30}{\milli\meter}
behind the array.
\item \emph{Hadamard (HAD)}: zero delays across elements with apodizations taken as rows of an order-80 Hadamard matrix~\cite{chiao1997sparse}.
\item \emph{Random $\pm 1$ (RND)}: i.i.d.\ Bernoulli $\pm 1$ apodizations with zero delays across elements. The incoherent compressed-sensing counterpart of
HAD.
\end{itemize}
During experiments, the acquisitions are subsampled with an interleaving pattern for the steered families, such that the entire sector remains covered.

\begin{figure*}[th]
    \centering
    \includegraphics[width=0.95\linewidth]{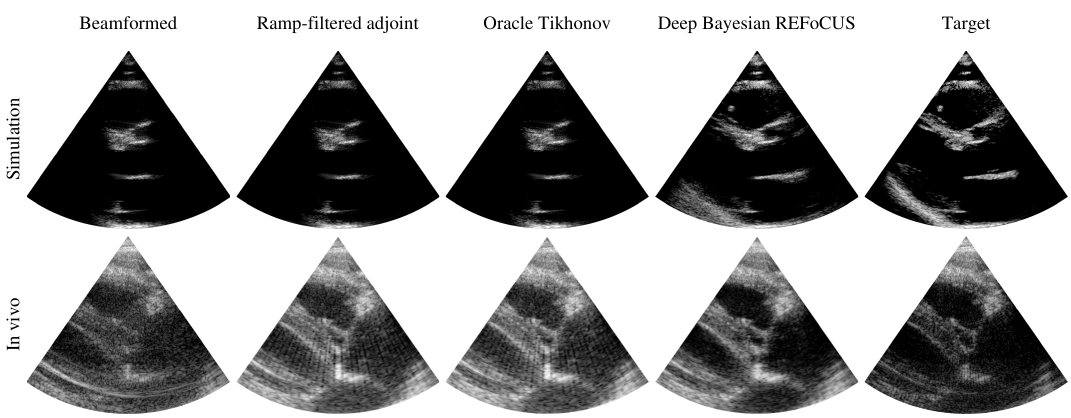}
    \caption{Comparison of reconstruction quality of linear REFoCUS decoders and Deep Bayesian REFoCUS. For the simulation a single plane wave is fired and the target is the simulation multistatic ground truth. For the in vivo 40 harmonic wide beams are fired, and the target shows 80 harmonic wide beams.}
    \label{fig:comparison}
\end{figure*}
\begin{figure}[!t]
    \centering
    \includegraphics[width=0.95\linewidth]{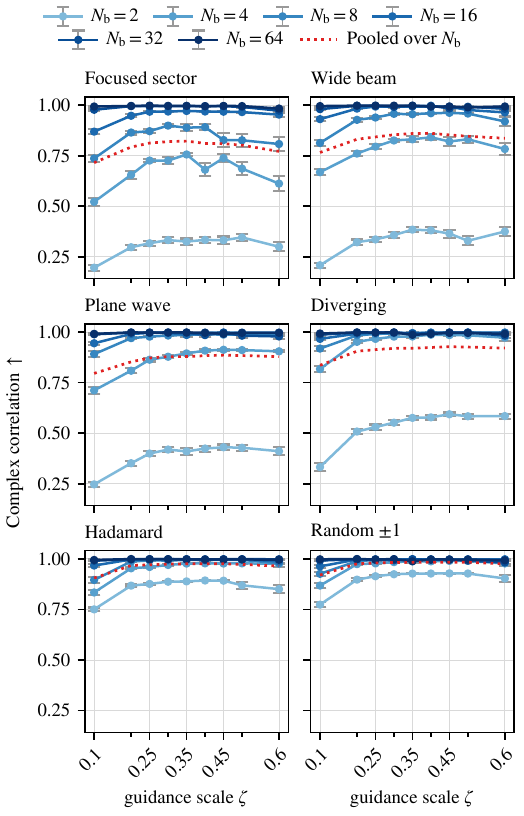}
    \caption{Complex correlation of DB-REFoCUS recovered multistatic data sets against the ground truth. Panels show different transmit schemes, the lines show different numbers of transmit events. We sweep guidance strength to find the global operating point $\zeta=0.35$.}
    \label{fig:ablation_zeta}
\end{figure}
\begin{figure}[!t]
    \centering
    \includegraphics[width=0.95\linewidth]{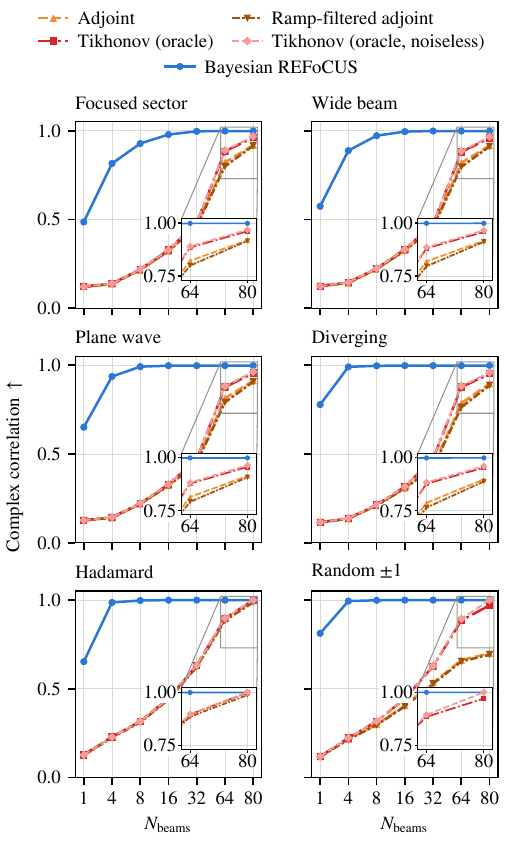}
    \caption{Complex correlation of DB-REFoCUS compared with the linear baselines for a range of $\Nb$ at \SI{10}{dB} measurement SNR. The noiseless Tikhonov method shows that linear recovery is limited by rank deficiency of the inversion.}
    \label{fig:e2plots}
\end{figure}

\subsection{Experiments}
\subsubsection{Simulation}
For the simulation experiments, the SNR profile of the in vivo acquisitions is copied by adding noise to the transmits with $\y = \A(\x) + \sigma G(z)\epsilon$. Here $\sigma$ is a scalar proportional to the target SNR, $z$ the depth, $G(z)$ the power-normalized TGC profile, and $\epsilon \sim \mathcal{CN}(0,1)$. 
Unless stated otherwise, we use a global measurement $\mathrm{SNR}=\SI{10}{dB}$ for all simulation experiments. The profile of $G(z)$ is defined across 32 blocks with \SI{2.951}{mm} spacing, spreading a global $\mathrm{SNR}$ of \SI{10}{dB} across local SNRs of $[35.95, 34.89, \dots, 1.99, 0.94]$\,\si{dB} with respect to depth.

First we sweep guidance strength $\zeta$ for 50 test frames for all transmit families and all $\Nb$ to select a single global operating point for DB-REFoCUS. Then, we evaluate multistatic recovery quality on 100 different test frames in the channel and image domains. To test noise robustness, we use $\Nb=[32, 80]$ and sweep $\mathrm{SNR} = [-20, -10, 0, 10, 20, \infty]$\,\si{dB} for the FOC and HAD transmit types. In this experiment we retune $\zeta$ on 10 separate test frames, select a static optimum per noise level, and evaluate on 100 separate frames. For the oracle Tikhonov method a grid search is still in place, and we select the best-performing result per frame.
Lastly, we calculate the Spearman correlation $\rho$ between the multistatic posterior variance and the multistatic loss in the channel domain for every transmit type at every $\Nb$. This experiment shows the impact of epistemic uncertainty, and whether the model can localize the rank deficiency of the measurements.

\subsubsection{In vivo}
For the in vivo experiments we use the Verasonics Vantage 256 system~\cite{daigle2012verasonics} with a Philips S5-1 probe with a \SI{0.254}{mm} pitch.
A non-expert operator records the cardiac PLAX view in the Left Lateral Recumbent (LLR) position for 12 anonymized volunteer subjects. For every transmit family we record 100 frames at \SI{33}{Hz} (\SI{3}{s}).

All acquisitions are transmitted and demodulated at $f_c=\SI{3.906}{MHz}$ and recorded with a sampling rate of $f_s=$\SI{15.625}{MHz}. For the FOC and WD families we extend the fundamental mode recordings with harmonic counterparts, which transmit at $f=$\SI{1.953}{MHz}. We apply a \SI{2}{MHz} wide band-pass filter, apply a 32-sample Hanning window, demodulate, and downsample by a factor of 3. The resulting IQ data is then shifted from \SI{4.1667}{MHz} to \SI{4}{MHz} with a phase ramp, the axial samples are cropped to the first 512 entries, and the recordings are normalized to unit variance with a single scalar. The image processing pipeline for the in vivo data is the same as that for the simulated data. 
The in vivo channel data set is made available as part of OpenH-RF~\cite{penninga_tue_cardiac_plax_2026}.

For the in vivo experiments we have no ground-truth multistatic data set, so we evaluate on held-out sets. For $\Nb = \{8, 16, 32, 64\}$ the multistatic data set is recovered with every method, transmit beamformed to the remaining $\{72, 64, 48, 16\}$ transmits, and compared with the real acquisitions. In the image domain we compare the B-mode created from the recovered multistatic with the B-mode of the acquisition with $\Nb=80$.
Because the model is not trained on in vivo acquisitions, $\zeta$ is increased to 0.8 for stronger measurement guidance for in vivo experiments.
\section{Results}
\label{sec:results}

A visual comparison of the methods is shown in Fig.~\ref{fig:comparison}.
For the simulation example, a single plane wave is fired at a 0$^\circ$ angle and the target is the full multistatic data set. The linear decoders stay true to the measurement and shrink the null-space components to 0, while DB-REFoCUS out-paints the B-mode. The result differs from the target, but the model shows an understanding of the cardiac anatomy that aligns with such a measurement.
For the in vivo example, the methods receive as measurement 40 harmonic wide beams and have as a target the equivalent frame with 80 harmonic wide beams. Shown is their beamformed multistatic recovery.
Here, the classical recovery methods clearly show the transmit beamforming pattern of radial streaks, as the null space between them is not filled in. The DB-REFoCUS result does not have this problem. However, it shows a clear distribution shift towards EchoNet-LVH, artificially increasing contrast.

For supplementary (interactive) qualitative and video results, visit \href{https://swpenninga.github.io/deep-bayesian-refocus/}{https://swpenninga.github.io/deep-bayesian-refocus}.

The results for the $\zeta$-sweep are shown in Fig.~\ref{fig:ablation_zeta}.
For the simulation data, the results are insensitive to the guidance strength parameter across all transmit types and measurement ranks. The optimum $\zeta=0.35$ is selected for the remaining simulation experiments.

The recovery results in Fig.~\ref{fig:e2plots} show that the DB-REFoCUS multistatic recovery often reaches a complex correlation of 1 with as few as 4 out of 80 transmits.
A clear relation can be seen between the focusing strength and the DB-REFoCUS recovery at low ranks. For $\Nb=1$ the focused sector shows the worst performance, as it has the least coverage of the sector. The wide-beam, plane-wave and diverging-wave transmits show increased performance, as they insonify a larger area of the sector. In this setting, the DB-REFoCUS model is thus not limited by SNR, but by epistemic uncertainty.
\begin{figure}[!t]
    \centering
    \includegraphics[width=0.85\linewidth]{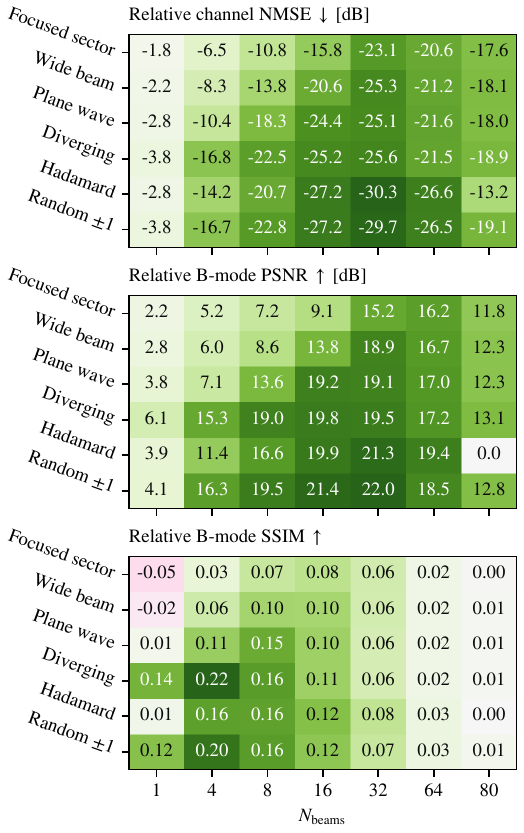}
    \caption{Heatmap showing NMSE, PSNR and SSIM of DB-REFoCUS relative to the best performing linear decoder, oracle Tikhonov.}
    \label{fig:e2heatmap}

    \vspace{\floatsep}
    \includegraphics[width=0.85\linewidth]{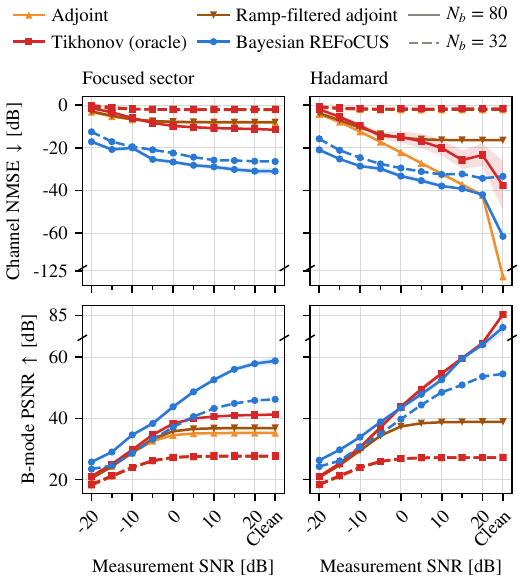}
    \caption{Performance of DB-REFoCUS compared with linear decoder methods for a range of SNR levels. The dashed lines show $\Nb=32$ performance, the solid lines show $\Nb=80$ performance.}
    \label{fig:noise}
\end{figure}
\begin{figure*}[!t]
    \begin{minipage}[t]{\columnwidth}
        \centering
        \includegraphics[width=\linewidth]{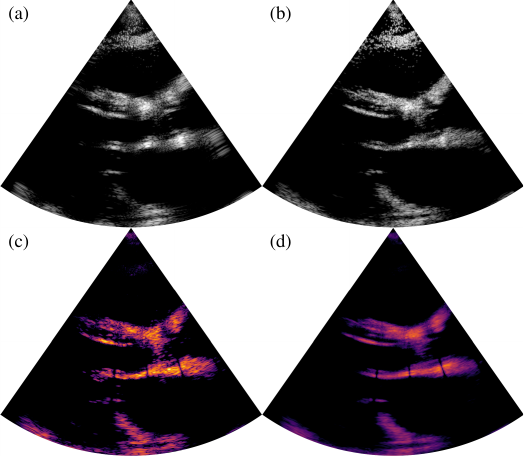}
        \caption{Posterior uncertainty and model loss shown for a DB-REFoCUS recovery. (a) A measurement of 8 focused beams, (b) model recovery, (c) loss with respect to the multistatic target, (d) posterior variance of 8 samples.}
        \label{fig:uncertainty}
    \end{minipage}\hfill
    \begin{minipage}[t]{\columnwidth}
        \centering
        \includegraphics[width=\linewidth]{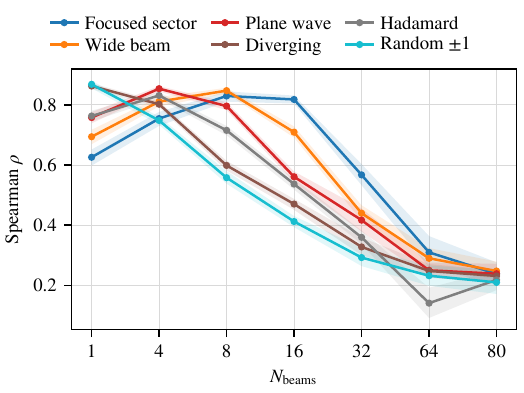}
        \caption{Spearman correlation coefficient $\rho$ between the DB-REFoCUS multistatic loss and the posterior variance, for a range of rank deficiencies and all transmit types.}
        \label{fig:correlation}
    \end{minipage}
\end{figure*}
Nevertheless, for every operating point DB-REFoCUS surpasses the linear decoders in performance. 
A noiseless oracle Tikhonov baseline is added to show that linear recovery is not bound by the noise, but by rank deficiency of the measurements. The noisy and noiseless Tikhonov baselines show similar performance for all $\Nb$.
\footnote{A steering angle of $45^\circ$ does not satisfy $d \leq \lambda / (1 + \sin\theta_{\max})$, so acquisitions at large angles contain grating lobes ($\theta_{\max} \approx33^\circ$). This is why the linear decoders do not reach a complex correlation of 1 for the steered transmit families. This does not affect the non-steered transmits or the in vivo harmonic imaging modes, which transmit at $f=$\SI{1.953}{MHz}.}
For every operating point the NMSE, PSNR and SSIM performance of DB-REFoCUS is shown relative to the oracle Tikhonov method in Fig.~\ref{fig:e2heatmap}. 
Both in the channel domain and image domain, DB-REFoCUS outperforms the Tikhonov baseline for all metrics. The difference is the strongest in the subsampling regime of $\Nb = [8, \dots, 32]$. Below this number of transmits, the model does not have enough information to fully recover the multistatic data set. DB-REFoCUS generates plausible samples, but the posterior spread is relatively large. The samples all align with the measurements, but this is not enough to align with the complete multistatic. For higher numbers of transmits there is less benefit in having a generative prior. The multistatic is better populated, and linear decoders perform relatively better in this regime, shrinking the performance gap.

To investigate the impact of noise on recovery, Fig.~\ref{fig:noise} shows an SNR sweep for the FOC and HAD transmit types for $\Nb = [32,80]$. 
The plots show that DB-REFoCUS outperforms the linear counterparts for all noise levels when $\Nb=32$. When removing the rank deficiency and comparing $\Nb=80$ we find that for low-SNR regimes DB-REFoCUS still performs better, but for a high enough SNR or clean image the linear decoders perform better. In this regime the linear inversion is exact, while the generative model still introduces stochasticity through sampling.

The relation between posterior uncertainty and rank deficiency is shown in Fig.~\ref{fig:uncertainty}.
The panels show posterior uncertainty and model loss for a measurement of 8 equispaced focused transmits. We sample 8 multistatic recoveries and beamform the residual of the Monte Carlo posterior mean with respect to the multistatic ground truth. Likewise, we beamform the variance of the multistatic samples and plot the result. Panels (c) and (d) agree. The top part of the reconstruction shows no variance because it is reconstructed correctly. The dark spots without tissue also show no variance or loss. Most notable are the focal points. Here the highest pressure was achieved and reconstruction of tissue is best. These radial streaks are present in both panels.

\begin{figure}[!t]
    \centering
    \includegraphics[width=\linewidth]{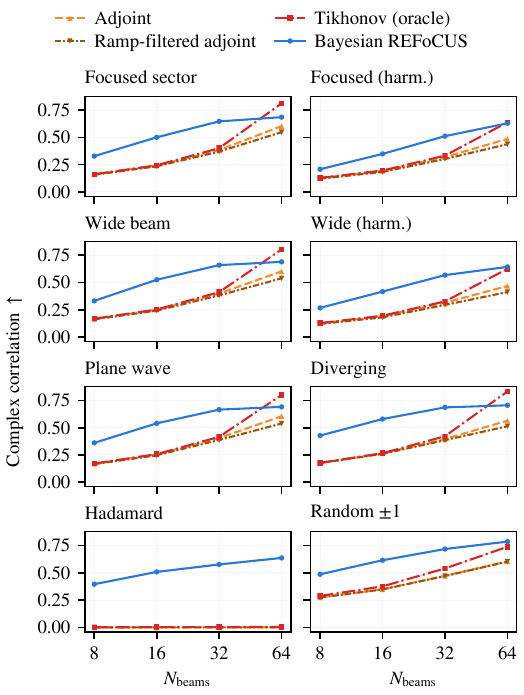}
    \caption{In vivo complex correlation for different $\Nb$ for all transmit types. The complex correlation is calculated on a held-out set of the remaining $80-\Nb$ transmits that belong to the same frame. The recovered multistatic data set is transmit beamformed to these acquisitions to compare.}
    \label{fig:invivo}
\end{figure}
\begin{figure}[!t]
    \centering
    \includegraphics[width=\linewidth]{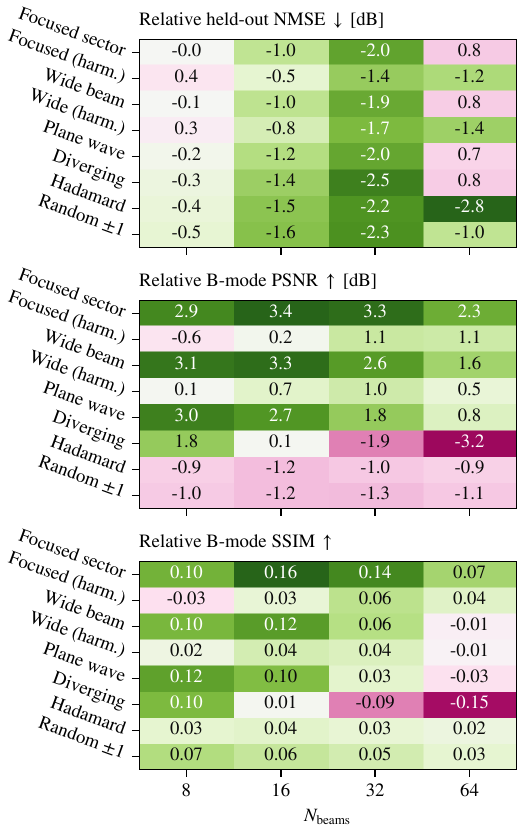}
    \caption{In vivo NMSE, PSNR and SSIM of DB-REFoCUS relative to the oracle Tikhonov linear decoder. NMSE is calculated through transmit beamforming to the held-out set, the image-domain metrics are calculated with respect to the $\Nb=80$ image.}
    \label{fig:invivoheatmap}
\end{figure}

The relation between the multistatic loss and posterior variance can be quantified with the Spearman correlation coefficient $\rho$. In Fig.~\ref{fig:correlation} a strong monotonic correlation is visible for high rank deficiencies across all transmit types.
As expected, for $\Nb\geq16$ the correlation drops sharply. The posterior converges to a single estimate as epistemic uncertainty is reduced. The correlation is now dominated by a fixed reconstruction gap and stochastic generation of samples.

The in vivo results for complex correlation are shown in Fig.~\ref{fig:invivo}. 
The model performs notably worse than for simulation data, due to a distribution gap. The model was not trained on this frequency, nor has it seen out-of-plane clutter, non-linear effects or physical 3D wave dispersion. Still, from simulation the model has learned coherence and the multistatic structure, allowing it to outperform the linear decoders for low-rank inversion. At higher counts for $\Nb$ the model pushes the measurements into the prior distribution, and this effect overtakes the gain of structuring the null space. Most notable is the performance of the Hadamard transmit type, where linear decoders fail by design. The Hadamard measurements are made orthogonal by design, such that unmeasured codes are fully unobserved. The DB-REFoCUS decoder still manages to recover a large component of the missing acquisitions through learned structure.

The other metrics are shown in Fig.~\ref{fig:invivoheatmap}. 
The NMSE result supports the complex correlation result, but the image-domain metrics deviate. This is because the EchoNet-LVH statistics are present in the recovery, as was shown in Fig.~\ref{fig:comparison}. We argue that out-of-distribution image-domain recovery is better evaluated visually, and refer to the URL mentioned at the start of this section.

\section{Discussion}
\label{sec:discussion}
The simulation results show the promising potential for this type of architecture to introduce deep generative models into the channel domain of ultrasound imaging. 

The main challenge that limits the method is the availability of high-quality multistatic data sets to train on. Such data is not prevalent and it can be difficult to obtain for all anatomies. We expect that training the model through ambient diffusion~\cite{daras2023ambient} might offer a solution here, such that the model can learn in vivo multistatic structure from any transmit type. 
However, there is a trade-off that needs to be addressed when training with real data. As became clear in Fig.~\ref{fig:comparison}, out-of-distribution guidance is possible and can have positive effects. If the model has only seen idealized simulation statistics, it will naturally denoise, remove artifacts and correct for aberration during recovery. When these artifacts become part of the training data, the model could lose these capabilities or even introduce aberration and similar artifacts during generation. Conversely, if the in vivo distribution is too far from the prior, the model is at risk of hallucinating~\cite{aithal2024hallucinations}.

The model has shown impressive generalization capabilities, such as the ability to work with harmonic imaging and even a slight distribution shift. There are further generalization steps that should be made for a true universal prior, however. The current U-Net structure limits the domain in which the model can operate. We foresee that a transformer architecture (DiT~\cite{peebles2023dit}) can further generalize across probe types, dimensions in space, frequencies, bandwidth, depth, number of probe elements, and anatomies, by entering each of these features as a conditioning vector rather than being fixed by the network architecture. 

Furthermore, the current model is orders of magnitude slower and more expensive than the linear alternatives. The main limitation is the number of steps $\Gamma$ that is required to transport the noise initialization to the multistatic. This requires a future investigation into model distillation for fewer steps, architecture pruning, or better initialization of the transport. Initializing transport with a linear REFoCUS recovery is one such solution. More accurate likelihood estimators like CovarianceDPS~\cite{yismaw2025codps} can also reduce the number of necessary guidance steps, and methods like SeqDiff~\cite{stevens2024sequential} can enable further improvements for video applications.

Lastly, the ability of the model to produce uncertainty maps and work with arbitrary transmit designs ties the work of this paper to the cognitive ultrasound framework~\cite{vansloun2024cognitive}. An information-seeking agent can use such a generative prior to adapt transmit-receive sequences in a Bayesian reconstruction loop.
Recent works have shown agent capabilities that select focused transmits for echocardiography~\cite{vannierop2026patientadaptive}, choose subsampling patterns~\cite{nolan2025ads}, and amortize sampling cost across frames~\cite{stevens2024sequential}. These implementations all work on beamformed, envelope, or image-domain data, and are limited to a fixed set of transmits. This work alleviates these constraints for future agents.
\section{Conclusion}
\label{sec:conclusion}
We have introduced Deep Bayesian REFoCUS, a deep generative prior in the channel domain that can conditionally generate multistatic data sets and obtain posterior distributions for all transmit types. The method outperformed classical linear REFoCUS decoders across the board in simulation, and showed promising results for in vivo data. The model performs best for inversion tasks that are extremely rank-deficient and regresses toward linear decoder performance when inversion is exact. Despite in vivo measurements being out of training distribution, the model was able to structure the multistatic for sufficient image reconstruction.
Further improvements of the method can be enabled by large-scale in vivo data collection, further generalization of the architecture and computational speedups.

\bibliographystyle{IEEEtran}
\bibliography{refs}

\end{document}